\documentclass[letterpaper]{article} 
\usepackage{aaai18}  
\usepackage{times}  
\usepackage{helvet}  
\usepackage{courier}  
\usepackage{url}  
\usepackage{graphicx}  
\usepackage{booktabs}
\usepackage{amssymb,amsmath}
\usepackage{multirow}
\usepackage{threeparttable}
\begin{document}
%
\title{Smarnet: Teaching Machines to Read and Comprehend Like Human}
\author{
	Zheqian Chen$^*_{\bigtriangleup}$, Rongqin Yang$^*$,Bin Cao$^\ddagger$, Zhou Zhao$^\dag$,  Deng Cai$^*$\\
	$^*$State Key Lab of CAD$\&$CG, College of Computer Science, Zhejiang University, Hangzhou, China\\
	$^\dag$College of Computer Science, Zhejiang University, Hangzhou, China\\
	$^\ddagger$Eigen Technologies, Hangzhou, China\\
	$\{$zheqianchen, rongqin.yrq, dengcai$\}$@gmail.com, bincao@aidigger.com, zhaozhou@zju.edu.cn \\
}
\maketitle

\begin{abstract}
Machine Comprehension (MC) is a challenging task in Natural Language Processing field, which aims to guide the machine to comprehend a passage and answer the given question. Many existing approaches on MC task are suffering the inefficiency in some bottlenecks, such as insufficient lexical understanding, complex question-passage interaction, incorrect answer extraction and so on. In this paper, we address these problems from the viewpoint of how humans deal with reading tests in a scientific way. Specifically, we first propose a novel lexical gating mechanism to dynamically combine the words and characters representations. We then guide the machines to read in an interactive way with attention mechanism and memory network. Finally we add a checking layer to refine the answer for insurance. The extensive experiments on two popular datasets SQuAD and TriviaQA show that our method exceeds considerable performance than most state-of-the-art solutions at the time of submission. 
\end{abstract}

\section{Introduction}
Teaching machines to learn reading comprehension is one of the core tasks in NLP field. Recently machine comprehension task accumulates much concern among NLP researchers. We have witnessed significant progress since the release of large-scale datasets like SQuAD~\cite{rajpurkar2016squad}, MS-MARCO~\cite{Nguyen2016MS}, TriviaQA~\cite{Joshi2017TriviaQAAL}, CNN/Daily Mail~\cite{Hermann2015TeachingMT} and Children's Book Test~\cite{Hill2015The}. The essential problem of machine comprehension is to predict the correct answer referring to a given passage with relevant question. If a machine can obtain a good score from predicting the right answer, we can say the machine is capable of understanding the given context.

Many previous approaches~\cite{seo2016bidirectional}~\cite{Gong2017Ruminating}~\cite{Wang2017GatedSN} adopt attention mechanisms along with deep neural network tactics and pointer network to establish interactions between the question and the passage. The superiority of these frameworks are to enable questions focus on more relevant targeted areas within passages. Although these works have achieved promising performance for MC task, most of them still suffer from the inefficiency in three perspectives: (1) Comprehensive understanding on the lexical and linguistic level. (2) Complex interactions among questions and passages in a scientific reading procedure. (3) Precise answer refining over the passage.

For all the time through, we consider a philosophy question: What will people do when they are having a reading comprehension test? Recall how our teacher taught us may shed some light. As a student, we recite words with relevant properties such as part-of-speech tag, the synonyms, entity type and so on. In order to promote answer's accuracy, we iteratively and interactively read the question and the passage to locate the answer's boundary. Sometimes we will check the answer to ensure the refining accuracy. Here we draw a flow path to depict what on earth the scientific reading skills are in the Figure 1. As we see, basic word understanding, iterative reading interaction and attentive checking are crucial in order to guarantee the answer accuracy.
\begin{figure}[t]
	\centering
	\includegraphics[width=8cm]{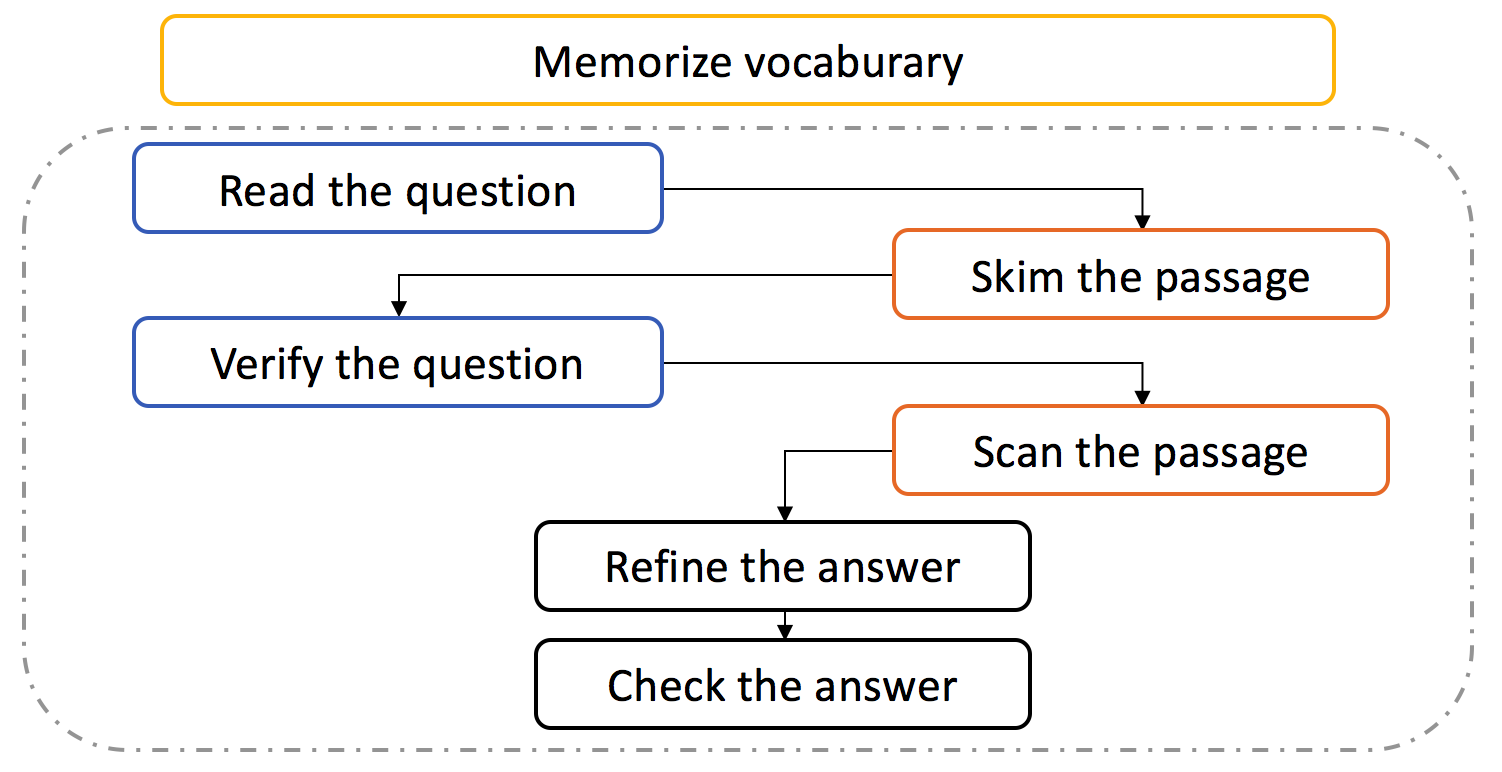}
	\caption{A scientific reading flow method}
\end{figure}
 \begin{figure*}[t]
	\centering
	\includegraphics[width=18cm]{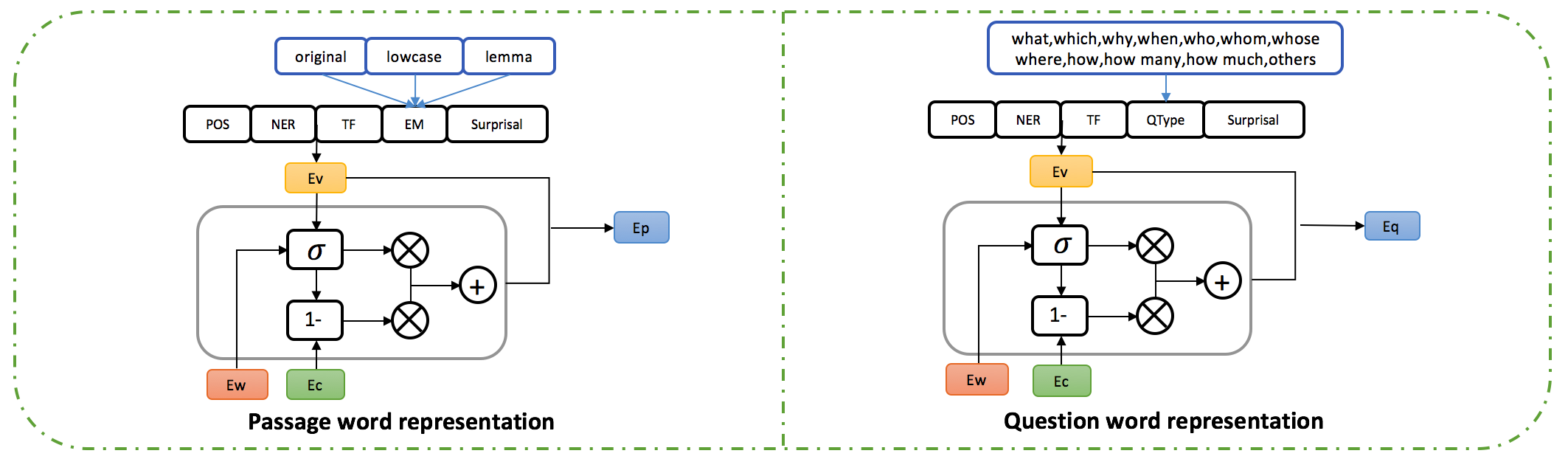}
	\caption{Fine-grained gating on lexical attributions of words and characters. ``POS, NER, TF, EM, Surprisal, QType'' refer to part-of-speech tags, named entity tags, term frequency, exact match, surprisal extent, question type.}
\end{figure*}

In this paper, we propose the novel framework named \textbf{Smarnet} with the hope that it can become as smart as humans. We design the structure in the view point of imitating how humans take the reading comprehension test. Specifically, we first introduce the \textbf{Smarnet} framework that exploits fine-grained word understanding with various attribution discriminations, like humans recite words with corresponding properties. We then develop the interactive attention with memory network to mimic human reading procedure. We also add a checking layer on the answer refining in order to ensure the accuracy. The main contributions of this paper are as follows:

\begin{itemize}
	\item We enrich the word representation with detailed lexical properties. We adopt a fine-grained gating mechanism to dynamically control the amount of word level and character level representations based on the properties of words.
	\item We guide the machine to read by imitating human's behavior in the reading procedure. We apply the Interactive attention to comprehensively model the question and passage representation. We adopt memory network to enhance the comprehending capacity and accuracy. 
	\item We utilize a checking mechanism with passage self alignment on the revised pointer network. This helps to locate the answer boundary and promote the prediction accuracy for insurance. 
\end{itemize}
 
 \section{Task Description}
The goal of open-domain MC task is to infer the proper answer from the given text. For notation, given a passage $P=\{p_1,p_2,...,p_m\}$ and a question $Q=\{q_1,q_2,...,q_n\}$, where $m$ and $n $ are the length of the passage and the question. Each token is denoted as $(w_i,C_i)$, where $w_i$ is the word embedding extracts from pre-trained word embedding lookups, $C_i$ is the char-level matrix representing one-hot encoding of characters. The model should read and comprehend the interactions between $P$ and $Q$, and predict an answer $A$ based on a continuous sub-span of $P$. 
 
 \section{ Smarnet Structure}
 The general framework of MC task can be coarsely summarized as a three-layer hierarchical process: Input embedding layer, Interaction modeling layer, answer refining layer. We then introduce our model from these three perspectives. 

 \subsection{Input Embedding Layer}
  Familiar with lexical and linguistic properties is crucial in text understanding. We try to enrich the lexical factors to enhance the word representation. Inspired by Yang et al.~\cite{Yang2016Words} \cite{Monsalve2012Lexical} \cite{Liu2017Structural} and \cite{chen2017reading}, we adopt a more fine-grained dynamic gating mechanism to model the lexical properties related to the question and the passage. Here we indicate our embedding method in Figure 2. We design two gates adopted as valves to dynamically control the flowing amount of word-level and character-level representations. 
  
   For the passage word $E_{vp}$, we use the concatenation of part-of-speech tag, named entity tag, term frequency tag, exact match tag and the surprisal tag. The exact match denote as $f_{em}(p_i)=\mathbb{I}(p_i \in q)$ in three binary forms: original, lower-case and lemma forms, which indicates whether  token $p_i$ in the passage can be exactly matched to a question word in $q$. The surprisal tag measures the amount of information conveyed by a particular word from  $Surprisal(w_t)=-log(P(w_t|w_1,w_2,...,w_{t-1}))$. The less occurrence of a word, the more information it carries. 
   
   For the question word $E_{vq}$, we take the question type in place of the exact match information and remain the other features. The type of a question provides significant clue for the answer selection process. For example, the answer for a "when" type question prefers tokens about time or dates while a "why" type question requires longer inference. Here we select the top 11 common question types as illustrated in the diagram. If the model recognize a question's type, then all the words in this question will be assigned with the same QType feature.
   
   The gates of the passage and the question are computed as follows:
   \begin{align*}
    g_p=\sigma(W_pE_{vp}+b_p) \\
   g_q=\sigma(W_qE_{vq}+b_q)   \tag{1} 
   \end{align*}
   where $W_p$, $b_p$, $W_q$, $b_q$ are the parameters and  $\sigma$denotes an element-wise sigmoid function.
   
   Using the fine-grained gating mechanism conditioned on the lexical features, we can accurately control the information flows between word-level and char-level. Intuitively, the formulation is as follows:
   \begin{align*}
   h_p=g(E_w,E_c)=g_p\circ E_w+(1-g_p)\circ E_c \\
   h_q=g(E_w,E_c)=g_q\circ E_w+(1-g_q)\circ E_c \tag{2}
   \end{align*}
   where $\circ$ is the element-wise multiplication operator. when the gate has high value, more information flows from the word-level representation; otherwise, char-level will take the dominating place. It is practical in real scenarios. For example, for unfamiliar noun entities, the gates tend to bias towards char-level representation in order to care richer morphological structure. Besides, we not only utilize the lexical properties as the gating feature, we also concatenate them as a supplement of lexical information. Therefore, the final representation of words are computed as follows:
   \begin{align*}
   E_p=[h_p,E_{vp}] \\
   E_q=[h_q,E_{vq}] \tag{3} 
   \end{align*} 
   where $[h,E]$ is the concatenation function.
 \subsection{Interaction Modeling Layer}
 Recall how people deal with reading comprehension test. When we get a reading test paper, we read the question first to have a preliminary focal point. Then we skim the passage to refine the answer. Sometimes we may not directly ensure the answer's boundary, we go back and confirm the question. After confirming, we scan the passage and refine the right answer we thought. We also check the answer for insurance. Inspired by such a scientific reading procedure, we design the \textbf{Smarnet} with three components: contextual encoding, interactive attention with memory network, answer refining with checking insurance. As is shown in figure 3.
 \begin{figure}[t]
 	\centering
 	\includegraphics[width=8cm]{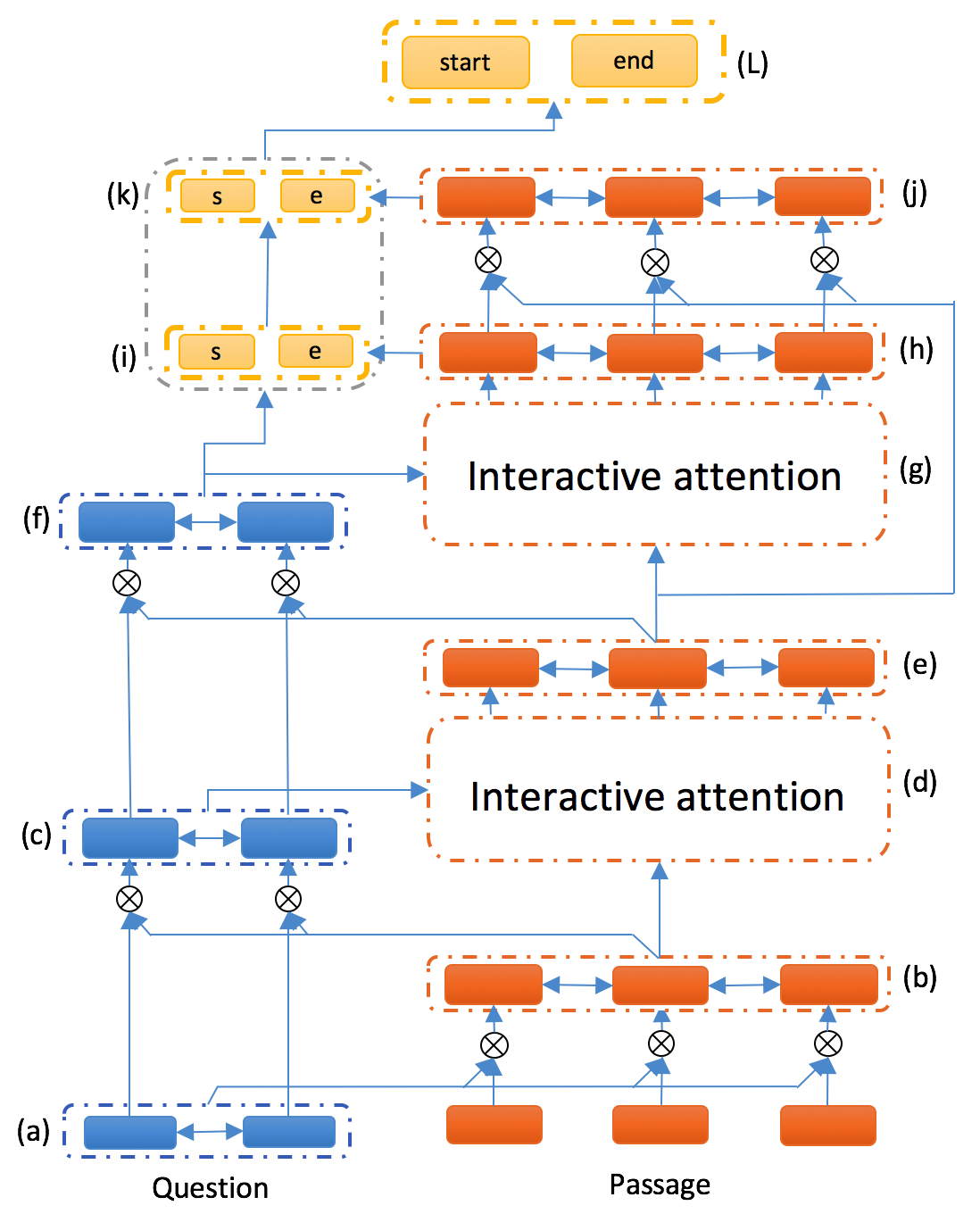}
 	\caption{The overview of \textbf{Smarnet} structure.}
 \end{figure}
 \subsubsection{Contextual Encoding} We use Gated Recurrent Unit~\cite{Cho2014Learning} with bi-directions to model the contextual representations. Here, It is rather remarkable that we do not immediately put the Bi-GRU on the passage words. Instead, we first encode the question and then apply a gate to control the question influence on each passage word, as is shown in the structure (a) and (b). Theoretically, when human do the reading comprehension, they often first read the question to have a rough view and then skim the passage with the impression of the question. No one can simultaneously read both the question and the passage without any overlap. Vice versa, after skimming the passage to get the preliminary comprehension, the whole passage representation is also applied to attend the question again with another gating mechanism, as is shown in the structure (c). This can be explained that people often reread the question to confirm whether they have thoroughly understand it. The outputs of the three steps (a) (b) (c) are calculated as follows:
 \begin{align*}
u^q_t &= BiGRU_q(u^q_{t-1}, E_t^q) \\
Q_1&= \left [ \overrightarrow{u_n^q},\overleftarrow{u_n^q} \right ] \tag{4}
 \end{align*} 
 where $E_t^q \in \mathbb{R}^d $ is the lexical representation from the input layer. $u_t^q  \in \mathbb{R}^d$ is the hidden state of GRU for the $t$th question word. $Q_1 \in \mathbb{R}^{2d}$ is the original question semantic embedding obtained from the concatenation of the last hidden states of two GRUs.  
 \begin{align*}
g_{q1}&=\sigma(W_{q1}Q_1+b_{q1})\\
E_{p1}&=g_{q1}\circ \beta Q_1+(1-g_{q1})\circ E_p\\
u_t^p&=GRU_p(u_{t-1}^p,E_t^{p1})\\
P_1&= \left [ \overrightarrow{u_m^p},\overleftarrow{u_m^p} \right ] \tag{5}
\end{align*}
where $g_{q1} \in \mathbb{R}^d$ is a question gate controlling the question influence on the passage. The larger the $g_{q1}$ is, the more impact the question takes on the passage word. We reduce the $Q_1$ dimension through multi-layer perceptron $\beta$ since $Q1$ and $E_{p1}$ are not in the same dimension. We then put the bi-GRU on top of each passage word to get the semantic representation of the whole passage $P_1$.    
\begin{align*}
g_{p1}&=\sigma(W_{p1}P_1+b_{p1})\\
E_{q2}&=g_{p1}\circ \beta P_1+(1-g_{p1}) \circ u_t^q \\
{u}'^q_t &= GRU_q({u}'^q_{t-1}, E_t^{q1}) \\
Q_2&= [ \overrightarrow{{u'}_n^q},\overleftarrow{{u'}_n^q}] \tag{6}
\end{align*}
 where $g_{p1}$ is a passage gate similar to $g_{q1}$. $\beta$ is the multi-layer perceptron to reduce dimension. $Q_2$ represents the confirmed question with the knowledge of the context.
 
 \subsubsection{Interactive Attention with Iterative Reasoning} 
 The essential point in answer refining lies on comprehensive understanding of the passage content under the guidance of the question. We build the interactive attention module to capture the mutual information between the question and the passage. From human's perspective, people repeatedly and interactively read the question and the passage, narrow the answer's boundary and put more attention on some passage parts where are more relevant to the question. 
 
 We construct a shared similarity matrix $S\in \mathbb{R}^{m\times n}$ to attend the relevance between the passage $P$ and the question $Q$. Each element $s_{ij}$ is computed by the similarity of the $i$th passage word and the $j$th question word. 

 We signify relevant question words into an attended question vector to collaborate with each context word. Let $a_j\in \mathbb{R}^n$ represent the normalized attention distribution on the question words by $t$th passage word. The attention weight is calculated by $a_j=softmax(S_{:j})\in \mathbb{R}^n$. Hence the attend question vector for all passage words $\widetilde{Q} \in \mathbb{R}^{2d}$ is obtained by $\widetilde{Q}=\sum_j a_j \cdot{Q_{:j}}$, where $Q_{:j} \in \mathbb{R}^{2d \times n}$.

We further utilize $\widetilde{Q}$ to form the question-aware passage representation. In order to comprehensively model the mutual information between the question and passage, we adopt a heuristic combining strategy to yield the extension as follows:
\begin{align*}
\widetilde{P_i^t}=BiGRU(\nu(h_i^q,\widetilde{Q},h_i^q \circ \widetilde{Q}, h_i^q + \widetilde{Q}) ) \tag{7}
\end{align*}
where $\widetilde{P_i^t}\in \mathbb{R}^{2d}$ denotes the $i$th question-aware passage word under the $t$th hop, the $\nu$ function is a concatenation function that fuses four input vectors. $h_i^q$ denotes the hidden state of former $i$th passage word obtained from BiGRU. $\circ$ denotes element-wise multiplication and $+$ denotes the element-wise plus. Notice that after the concatenation of $\nu$, the dimension can reaches to $8d$. We feed the concatenation vector into a BiGRU to reduce the hidden state dimension into $2d$. Using BiGRU to reduce dimension provides an efficient way to facilitate the passage semantic information and enable later semantic parsing.

Naive one-hop comprehension may fail to comprehensively understand the mutual question-passage information. Therefore, we propose a multi-hop memory network which allows to reread the question and answer. In our model, we totally apply two-hops memory network, as is depicted in the structure (c to e) and (f to h). In our experiment we found the two-hops can reach the best performance. In detail, the memory vector stores the question-aware passage representations, the old memory's output is updated through a repeated interaction attention. 
\subsection{Answer Selection with Checking Mechanism}
The goal of open-domain MC task is to refine the sub-phrase from the passage as the final answer. Usually the answer span $(i, j)$ is derived by predicting the start $p_s^i$ and the end $p_e(j|i)$ indices of the phrase in the paragraph. In our model, we use two answer refining strategies from different levels of linguistic understanding: one is from original interaction output module and the other is from self-matching alignment. The two extracted answers are then applied into a checking component to final ensure the decision.

For the original interaction output in structure (h), we directly aggregate the passage vectors through BiGRU. We compute the $p_s^i$ and $p_e(j|i)$ probability distribution under the instruction of~\cite{Wang2016Machine} and pointer network~\cite{Vinyals2015PointerN} by  
\begin{align*}
p_{s1}=softmax(w^T_{s1}\widetilde{P_i^1})\\
p_{e1}=softmax(w^T_{e1}\widetilde{P_i^1}) \tag{8}
\end{align*}
where $\widetilde{P_i}$ is the output of the original interaction. $w^T_s \in \mathbb{R}^{2d}$ and $w^T_e \in \mathbb{R}^{2d}$ are trainable weight vectors.  

For the self-alignment in structure (j), we align the two hops outputs $\widetilde{P_{i1}}$ with $\widetilde{P_{i2}}$ in structure (e) and (h). The purpose of self-alignment aims to analysis the new insights on the passage as the comprehension gradually become clear after iterative hops. For each hop, the reader dynamically collects evidence from former passage representation and encodes the evidence to the new iteration. From human's perspective, each time we reread the passage, we get some new ideas or more solid comprehension on the basis of the former understanding. The self-alignment is computed by
\begin{align*}
\widetilde{P_{1}}&= \widetilde{P_{n}^1} \\
g_{\widetilde{P_{i1}}}&=\sigma(W_{\widetilde{P_{1}}}\widetilde{P_{1}}+b_{\widetilde{P_{1}}})\\
E_{\widetilde{P_{i2}}}&=g_{\widetilde{P_{i1}}}\circ  \widetilde{P_{1}}+(1-g_{\widetilde{P_{i1}}})\circ \widetilde{P_i^2}\\
\widetilde{{P_i}^2}&=BiGRU(\widetilde{P_{i-1}^2},E_{\widetilde{P_{i2}}}) \tag{9}
\end{align*} 
where $\widetilde{P_{1}}\in \mathbb{R}^{2d}$ is the first hop whole passage vector in the structure (e). We apply a gate mechanism with $\widetilde{P_{1}}$ to control the evidence flow to the next hop $\widetilde{P_{i2}} \in \mathbb{R}^{2d}$. The output of self-alignment is computed by
\begin{align*}
p_{s2}=softmax(w^T_{s2}\widetilde{P_i^2})\\
p_{e2}=softmax(w^T_{e2}\widetilde{P_i^2}) \tag{10}
\end{align*}
 where $p_{s2}$ and $p_{e2}$ are the predicted start and end indices after the self-alignment.
 
 For insurance, we obtain two groups of predicted answer span $P_{s1}\sim P_{e1}$ and $P_{s2}\sim P_{e2}$. We then apply a checking strategy to compare the twice answer. This process is quite similar to human's behavior, people often reread the passage and may draw some different answers. Thus they need to compare the alternative answers and finally consider a best one. Here we employ the weighted sum of twice answer prediction indices to make the final decision:
 \begin{align*}
 p_{s}&=p_{s1}+\alpha p_{s2}\\
 p_{e}&=p_{e1}+\alpha p_{e2}\\
 s&=argmax(p_s^1,p_s^2,...,p_s^m) \\
 e&=argmax(p_e^1,p_e^2,...,p_e^m) \tag{11}
 \end{align*} 
 where $\alpha \geq 1$ is a weighted scalar controlling the proportion of the two predicted answers. We set the $\alpha \geq 1$ as in most cases the latter predicted answer is more accurate comparing to the former one. The final $s$ and $e$ is then judged by the max value through the argmax operator.  
 
\subsection{Training and Optimization}
We choose the training loss as the sum of the negative log probabilities of the true start and end position by the predicted distributions to train our model:
\begin{align}
L(\Theta)=-\frac{1}{N}\sum_{i}^{N}logp_s^L(y_i^s)+logp_e^L(y_i^e) \tag{12}
\end{align}
where $\Theta$ denotes all the model coefficients including the neural network parameters and the input gating function parameters, N is the number dataset examples, $p_s^L$ and $p_e^L$ are the predicted distributions of the output, $y_i^s$ and $y_i^e$ are the true start and end indices of the $i$th example. The objective function in our learning process is given by:
\begin{align}
min\pounds  (\Theta)=L(\Theta)+\lambda \left \| \Theta \right \|^2 \tag{13}
\end{align}
where $\lambda$ is the trade-off parameter between the training loss and regularization. To optimize the objective, we employ the stochastic gradient descent (SGD) with the diagonal variant of AdaDelta~\cite{Zeiler2012ADADELTA}.

 \section{Experiments}
 \subsection{Datasets}
 In this section we evaluate our model on the task of machine comprehension using the recently released large-scale datasets SQuAD~\cite{rajpurkar2016squad} and TriviaQA~\cite{Joshi2017TriviaQAAL}. SQuAD published by Stanford has obtained a huge attention over the past two years. It is composed of over 100K questions manually annotated by crowd workers on 536 Wikipedia articles. TriviaQA is a newly released open-domain MC dataset which consists of over 650K question-answer-evidence triples. It is derived by combining 95K Trivia enthusiast authored question-answer pairs with on average six supporting evidence documents per question. The length of contexts in TriviaQA is much longer than SQuAD and models trained on TriviaQA require more cross sentence reasoning to find answers. 
 
 There are some similar settings between these two datasets. Each answer to the question is a segment of text from the corresponding reading context. Two metrics are used to evaluate models: Exact Match (EM) measures the percentage of predictions that match the ground truth answer exactly. F1 score measures the average overlap between the prediction and ground truth answer. Both datasets are randomly partitioned into training set (80\%), dev set (10\%) and test set (10\%). 
 
 \subsection{Implemental Details}
 We preprocess each passage and question using the library of nltk~\cite{Loper2002NLTKTN} and exploit the popular pre-trained word embedding GloVe with 100-dimensional vectors~\cite{Pennington2014GloveGV} for both questions and passages. The size of char-level embedding is also set as 100-dimensional and is obtained by CNN filters under the instruction of \cite{Kim2014ConvolutionalNN}. The Gated Recurrent Unit~\cite{Cho2014Learning} which is variant from LSTM~\cite{Hochreiter1997Long} is employed throughout our model. We adopt the AdaDelta~\cite{Zeiler2012ADADELTA} optimizer for training with an initial learning rate of 0.0005. The batch size is set to be 48 for both the SQuAD and TriviaQA datasets. We also apply dropout~\cite{Srivastava2014Dropout} between layers with a dropout rate of 0.2. For the multi-hop reasoning, we set the number of hops as 2 which is imitating human reading procedure on skimming and scanning. During training, we set the moving averages of all weights as the exponential decay rate of 0.999~\cite{Lucas2000ExponentiallyWM}. The whole training process takes approximately 14 hours on a single 1080Ti GPU. Furthermore, as the SQuAD and TriviaQA are competitive MC benchmark, we train an ensemble model consisting of 16 training runs with the same architecture but identical hyper-parameters. The answer with the highest sum of the confidence score is chosen for each query.
 
 \subsection{Overall Results}
 \begin{table}[!t]
	\centering
	\begin{tabular}{lclcl}
		\toprule
		Model & EM &F1 \\
		\midrule
		Match-LSTM with Bi-Ans-Ptr (single) &64.744&73.743\\
		DCN (single) &66.233&75.896\\
		BIDAF (single) &67.974 &77.323 \\
		SEDT(single) &68.163 &77.527\\
		RaSoR (single) &70.849 &78.741\\
		Multi-Perspective Matching (single) &70.387 &78.784 \\
		FastQAExt (single) &70.849 &78.857\\
		Document Reader (single) &70.333 &79.353 \\
		ReasoNet (single) &70.555 &79.364 \\
		Ruminating Reader (single) &70.639 &79.456 \\
		jNet (single) &70.607 &79.821 \\
		Mnemonic Reader (single) &70.995 &80.146 \\
		\textbf{Smarnet} (single) &\textbf{71.415} &\textbf{80.160} \\
		r-net (single) &74.614 &82.458 \\
		\midrule
		Human Performance &82.304        &91.221  \\
		\bottomrule	
	\end{tabular}
	\caption{Performance of single \textbf{Smarnet} model against other strong competitors on the SQuAD. The results are recorded on the submission date on July 14th, 2017.}
\end{table}
 \begin{table}[!t]
	\centering
	\begin{tabular}{lclcl}
		\toprule
		Model & EM &F1 \\
		\midrule
		Match-LSTM with Ans-Ptr (ensemble) &67.901 &77.022 \\
		DCN (ensemble) &71.625 &80.383 \\
		Multi-Perspective Matching (ensemble) &73.765 &81.257 \\
		jNet (ensemble) &73.010 &81.517 \\
	    BIDAF (ensemble) &73.744 &81.525 \\
		SEDT (ensemble) &74.090 &81.761 \\
		Mnemonic Reader (ensemble) &74.268 &82.371 \\
		ReasoNet (ensemble) &75.034 &82.552 \\
		MEMEN (ensemble) &75.370 &82.658 \\
		\textbf{Smarnet} (ensemble) &\textbf{75.989} &\textbf{83.475} \\
		r-net (ensemble) &77.688 &84.666\\
		\midrule
		Human Performance &82.304        &91.221  \\
		\bottomrule	
	\end{tabular}
\caption{Performance of ensemble \textbf{Smarnet} model against other strong competitors on the SQuAD. The results are recorded on the submission date on July 14th, 2017.}
\end{table}
\begin{table}[!t]
\centering
\begin{tabular}{c|c|c|c|c|clcl}
	\hline
	\multirow{2}{*}{Model} &\multirow{2}{*}{Domain} & \multicolumn{2}{|c|}{Full} & \multicolumn{2}{|c}{Verified} \\
	\cline{3-6}
	&& EM & F1  & EM & F1 \\
	\hline
	Random &\multirow{4}{*}{Wiki}& 12.74& 22.35 & 15.41 & 25.44 \\
	Classifier && 22.45 & 26.52 & 27.23 & 31.37 \\
	BiDAF && 40.32 & 45.91 & 44.86 & 50.71 \\
	\textbf{Smarnet} && \textbf{42.41} & \textbf{48.84} & \textbf{50.51} & \textbf{55.90} \\
	\hline
	Classifier &\multirow{3}{*}{Web}& 24.00 & 28.38 & 30.17 & 34.67 \\
	BiDAF && 40.74 & 47.05 & 49.54 & 55.80 \\
	\textbf{Smarnet} && \textbf{40.87} & \textbf{47.09} & \textbf{51.11} & \textbf{55.98} \\
	\hline
\end{tabular}
\caption{Performance of single \textbf{Smarnet} model against other strong competitors on the TriviaQA. The results are recorded on the submission date on September 3th, 2017.}
\end{table}

We evaluate the performance of our proposed method based on two evaluation criteria EM and F1 for the MC tasks. We compare our model with other strong competitive methods on the SQuAD leaderboard and TriviaQA leaderboard.

Table 1 and Table 2 respectively show the performance of single and ensemble models on the SQuAD leaderboard. The SQuAD leaderboard is very competitive among top NLP researchers from all over the world. We can see the top record has been frequently broken in order to reach the human's level. Our model was submitted by July 14, 2017, thus we compare our model on the single and ensemble performance against other competitor at that time.

From the tables 1 and 2 we can see our single model achieves an EM score of $71.415\%$ and a F1 score of $80.160\%$ and the ensemble model improves to EM $75.989\%$ and F1 $83.475\%$, which are both only after the r-net method~\cite{Wang2017GatedSN} at the time of submission. These results sufficiently prove the significant superiority of our proposed model.

We also compare our models on the recently proposed dataset TriviaQA. Table 3 shows the performance comparison on the test set of TriviaQA. We can see our \textbf{Smarnet} model outperforms the other baselines on both wikipedia domain and web domain.
\subsection{Ablation Results}
\begin{table}[!t]
	\centering
	\begin{tabular}{lclcl}
		\toprule
		Features & EM &F1 \\
		\midrule
		Full & \textbf{71.362}        &\textbf{80.183}  \\
		\midrule
		No $f_{pos}$ &68.632      &78.911  \\
		No $f_{ner}$ &70.804        &79.257\\
		No $f_{em}$ &68.486       &78.589\\
		No $f_{surprisal}$ &70.972&79.021\\
		No $f_{tf}$ &70.617           &79.701\\
		No $f_{Q_{type}}$ &68.913&78.151\\
		No $f_{pos}$ and $ f_{ner}$ &67.352&77.239\\
		\bottomrule	
	\end{tabular}
	\caption{Lexical feature ablations on SQuAD dev set.}
\end{table}
\begin{table}[!t]
	\centering
	\begin{tabular}{lclcl}
		\toprule
		Model & EM &F1 \\
		\midrule
		Full & \textbf{71.362}        &\textbf{80.183}  \\
		\midrule
		Input concatenation &69.077  &78.531  \\
		Passage direct encoding &70.012&78.907 \\
		Memory network &68.774&77.323\\
		Self-alignment checking&69.395& 79.227 \\
		\bottomrule	
	\end{tabular}
	\caption{Component ablations on SQuAD dev set.}
\end{table}
\begin{figure}[t]
	\centering
	\includegraphics[width=8cm]{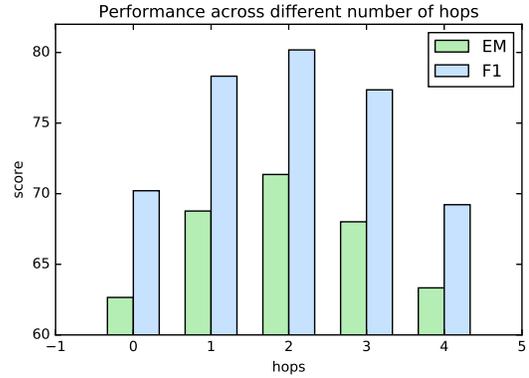}
	\caption{Performance across different number of hops}
\end{figure}
\begin{figure}[t]
	\centering
	\includegraphics[width=8cm]{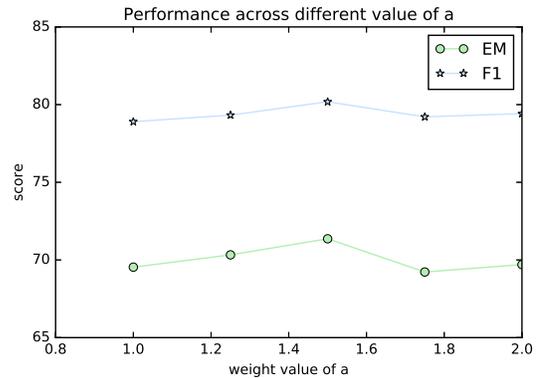}
	\caption{Performance across different weight of $\alpha$}
\end{figure}
We respectively evaluate the individual contribution of the proposed module in our model. We conduct thorough ablation experiments on the SQuAD dev set, which are recorded on the table 4 and table 5. 

Table 4 shows the different effect of the lexical features. We see the full features integration obtain the best performance, which demonstrates the necessity of combining all the features into consideration. Among all the feature ablations, the Part-Of-Speech, Exact Match, Qtype features drop much more than the other features, which shows the importance of these three features. As the POS tag provides the critical lexical information, Exact Match and Qtype help to guide the attention in the interaction procedure. As for the final ablation of POS and NER, we can see the performance decays over 3\% point, which clearly proves the usefulness of the comprehensive lexical information.   

Table 5 shows the ablation results on the different levels of components. We first replace our input gate mechanism into simplified feature concatenation strategy, the performance drops nearly 2.3\% on the EM score, which proves the effectiveness of our proposed dynamic input gating mechanism. We then compare two methods which directly encode the passage words or use the question influence. The result proves that our modification of employing question influence on the passage encoding can boost the result up to 1.3\% on the EM score. In our model, we apply two-hops memory network to further comprehend the passage. In the ablation test, we remove the iterative hops of memory network and only remain one interaction round. The result drops 2.6\% point on the EM score, which indicate the significance of using memory network mechanism. Finally, we compare the last module of our proposed self-alignment checking with original pointer network. The final result shows the superiority of our proposed method.

\subsection{Parameters Tuning}
We conduct two parameters tuning experiments in order to get the optimal performance. Figure 4 shows the results on different hops of memory network. We see the number of hops set to 2 can get the best performance comparing to other number of hops. In addition, as the number of hops enlarges, the model is easily to get overfitting on the training set, hence the performance is decrease rather than increase. In figure 5, we set different weight of $\alpha$ into five groups $\left\{1.0,1.25,1.5,1.75,2.0\right\}$. The final results show that the proportion of the first answer prediction and the last one reaches to 2:3 can get the most confident answer judgement. The value of $\alpha$ which is greater than 1, indicating that the latter answer refining takes more insurance on the prediction decision.

 \section{Related Work}
 \noindent\textbf{Machine Comprehension Dataset.} Benchmark datasets play a vital role in the research advance. Previous human-labeled datasets on MC task are too small to train data-intensive models~\cite{Richardson2013MCTestAC}~\cite{Berant2014ModelingBP}~\cite{Yu2016End}. Recently, Large-scale datasets become available. CNN/Daily Mail~\cite{Hermann2015TeachingMT} and Children's Book Test~\cite{Hill2015The} generated in cloze style offer the availability to train more expressive neural models. The SQuAD~\cite{rajpurkar2016squad}, TriviaQA~\cite{Joshi2017TriviaQAAL} and MS-MARCO~\cite{Nguyen2016MS} datasets provide large and high-quality datasets which extract answers from text spans instead of single entities in cloze style. The open-domain style of MC tasks are more challenging and require different levels of reasoning from multiple sentences. In this paper, we evaluate our \textbf{Smarnet} framework on SQuAD and TriviaQA datasets. 
 
 \noindent \textbf{Machine Comprehension Models} Previous works in MC task adopt deep neural modeling strategies with attention mechanisms, both on cloze style and open domain tasks. Along with cloze style datasets, Chen et al.~\cite{Chen2016A} prove that computing the attention weights with a bilinear term instead of simple dot-product significantly improves the accuracy. Kadlec et al.~\cite{Kadlec2016Text} sum attention over candidate answer words in the document. Dhingra et al.~\cite{Dhingra2017Gated} iteratively interact between the query and document by a multiplicative gating function. Cui et al.~\cite{Cui2016Attention} compute a similarity matrix with two-way attention between the query and passage mutually. Sordoni et al.~\cite{Sordoni2016Iterative} exploit an iterative alternating neural attention to model the connections between the question and the passage. 
 
 Open-domain machine comprehension tasks are more challenging and have attracted plenty of teams to pursue for higher performance on the leaderboard. Wang et al.~\cite{Wang2016Machine} present match-LSTM and use pointer network to generate answers from the passage. Chen et al.~\cite{chen2017reading} tackle the problem by using wikipedia as the unique knowledge source. Shen~\cite{shen2017reasonet} adopt memory networks with reinforcement learning so as to dynamically control the number of hops. Seo et al. \cite{seo2016bidirectional} use bi-directional attention flow mechanism and a multi-stage hierarchical process to represent the context. Xiong et al.~\cite{xiong2016dynamic} propose dynamic coattention networks to iteratively infer the answer. Yang et al.~\cite{Yang2016Words} present a fine-grained gating mechanism to dynamically combine word-level and character-level representations. Wang et al.~\cite{Wang2017GatedSN} introduce self-matching attention to refine the gated representation by aligning the passage against itself. 
 
 \noindent \textbf{Reasoning by Memory Network} Multi-hop reasoning combines with Memory networks have shown powerful competence on MC task~\cite{shen2017reasonet} \cite{Dhingra2017Gated} \cite{Sordoni2016Iterative} \cite{xiong2016dynamic} \cite{Hu2017Mnemonic} \cite{Gong2017Ruminating} \cite{Kumar2016AskMA}. Theoretically, multi-hop memory networks can repeat computing the attention biases between the query and the context through multiple layers. The memory networks typically maintain memory states which incorporate the information of current reasoning with the previous storage in the memory. Hu et al.~\cite{Hu2017Mnemonic} utilize a multi-hop answer pointer which allows the network to continue refining the predicted answer span. Gong et al.~\cite{Gong2017Ruminating} adapt the BIDAF~\cite{seo2016bidirectional} with multi-hop attention mechanisms and achieve substantial performance. Pan et al.~\cite{Pan2017MEMEN} introduce multi-layer embedding with memory network for full orientation matching on MC task. In our paper, we also adopt the memory network to mimic human behaviors on increasing their understanding by reread the context and the query multi times. We also apply a multi-hop checking mechanism to better refine the true answer.

\section{Conclusions}
In this paper, we tackle the problem of machine comprehension from the viewpoint of imitating human's ways in having reading comprehension examinations. We propose the \textbf{Smarnet} framework with the hope that it can become as smart as human for the reading comprehension problem. We first introduce a novel gating method with detailed word attributions to fully exploit prior knowledge of word semantic understanding. We then adopt a scientific procedure to guide machines to read and comprehend by using interactive attention and matching mechanisms between questions and passages. Furthermore, we employ the self-alignment with checking strategy to ensure the answer is refined after careful consideration. We evaluate the performance of our method on two large-scale datasets SQuAD and TriviaQA. The extensive experiments demonstrate the superiority of our \textbf{Smarnet} framework. 
\bibliographystyle{aaai}
\bibliography{formatting-instructions-latex-2018}

\begin{thebibliography}{}

\bibitem[\protect\citeauthoryear{Berant \bgroup et al\mbox.\egroup
  }{2014}]{Berant2014ModelingBP}
Berant, J.; Srikumar, V.; Chen, P.-C.; Linden, A.~V.; Harding, B.; Huang, B.;
  Clark, P.; and Manning, C.~D.
\newblock 2014.
\newblock Modeling biological processes for reading comprehension.
\newblock In {\em EMNLP}.

\bibitem[\protect\citeauthoryear{Chen \bgroup et al\mbox.\egroup
  }{2017}]{chen2017reading}
Chen, D.; Fisch, A.; Weston, J.; and Bordes, A.
\newblock 2017.
\newblock Reading wikipedia to answer open-domain questions.
\newblock {\em arXiv preprint arXiv:1704.00051}.

\bibitem[\protect\citeauthoryear{Chen, Bolton, and Manning}{2016}]{Chen2016A}
Chen, D.; Bolton, J.; and Manning, C.~D.
\newblock 2016.
\newblock A thorough examination of the cnn/daily mail reading comprehension
  task.

\bibitem[\protect\citeauthoryear{Cho \bgroup et al\mbox.\egroup
  }{2014}]{Cho2014Learning}
Cho, K.; Merrienboer, B.~V.; Gulcehre, C.; Bahdanau, D.; Bougares, F.; Schwenk,
  H.; and Bengio, Y.
\newblock 2014.
\newblock Learning phrase representations using rnn encoder-decoder for
  statistical machine translation.
\newblock {\em Computer Science}.

\bibitem[\protect\citeauthoryear{Cui \bgroup et al\mbox.\egroup
  }{2016}]{Cui2016Attention}
Cui, Y.; Chen, Z.; Wei, S.; Wang, S.; Liu, T.; and Hu, G.
\newblock 2016.
\newblock Attention-over-attention neural networks for reading comprehension.

\bibitem[\protect\citeauthoryear{Dhingra \bgroup et al\mbox.\egroup
  }{2017}]{Dhingra2017Gated}
Dhingra, B.; Liu, H.; Yang, Z.; Cohen, W.~W.; and Salakhutdinov, R.
\newblock 2017.
\newblock Gated-attention readers for text comprehension.

\bibitem[\protect\citeauthoryear{Gong and Bowman}{2017}]{Gong2017Ruminating}
Gong, Y., and Bowman, S.~R.
\newblock 2017.
\newblock Ruminating reader: Reasoning with gated multi-hop attention.

\bibitem[\protect\citeauthoryear{Hermann \bgroup et al\mbox.\egroup
  }{2015}]{Hermann2015TeachingMT}
Hermann, K.~M.; Kocisk{\'y}, T.; Grefenstette, E.; Espeholt, L.; Kay, W.;
  Suleyman, M.; and Blunsom, P.
\newblock 2015.
\newblock Teaching machines to read and comprehend.
\newblock In {\em NIPS}.

\bibitem[\protect\citeauthoryear{Hill \bgroup et al\mbox.\egroup
  }{2015}]{Hill2015The}
Hill, F.; Bordes, A.; Chopra, S.; and Weston, J.
\newblock 2015.
\newblock The goldilocks principle: Reading children's books with explicit
  memory representations.
\newblock {\em Computer Science}.

\bibitem[\protect\citeauthoryear{Hochreiter and
  Schmidhuber}{1997}]{Hochreiter1997Long}
Hochreiter, S., and Schmidhuber, J.
\newblock 1997.
\newblock Long short-term memory.
\newblock {\em Neural Computation} 9(8):1735--1780.

\bibitem[\protect\citeauthoryear{Hu, Peng, and Qiu}{2017}]{Hu2017Mnemonic}
Hu, M.; Peng, Y.; and Qiu, X.
\newblock 2017.
\newblock Mnemonic reader: Machine comprehension with iterative aligning and
  multi-hop answer pointing.

\bibitem[\protect\citeauthoryear{Joshi \bgroup et al\mbox.\egroup
  }{2017}]{Joshi2017TriviaQAAL}
Joshi, M.; Choi, E.; Weld, D.~S.; and Zettlemoyer, L.~S.
\newblock 2017.
\newblock Triviaqa: A large scale distantly supervised challenge dataset for
  reading comprehension.
\newblock In {\em ACL}.

\bibitem[\protect\citeauthoryear{Kadlec \bgroup et al\mbox.\egroup
  }{2016}]{Kadlec2016Text}
Kadlec, R.; Schmid, M.; Bajgar, O.; and Kleindienst, J.
\newblock 2016.
\newblock Text understanding with the attention sum reader network.
\newblock  908--918.

\bibitem[\protect\citeauthoryear{Kim}{2014}]{Kim2014ConvolutionalNN}
Kim, Y.
\newblock 2014.
\newblock Convolutional neural networks for sentence classification.
\newblock In {\em EMNLP}.

\bibitem[\protect\citeauthoryear{Kumar \bgroup et al\mbox.\egroup
  }{2016}]{Kumar2016AskMA}
Kumar, A.; Irsoy, O.; Ondruska, P.; Iyyer, M.; Bradbury, J.; Gulrajani, I.;
  Zhong, V.; Paulus, R.; and Socher, R.
\newblock 2016.
\newblock Ask me anything: Dynamic memory networks for natural language
  processing.
\newblock In {\em ICML}.

\bibitem[\protect\citeauthoryear{Liu \bgroup et al\mbox.\egroup
  }{2017}]{Liu2017Structural}
Liu, R.; Hu, J.; Wei, W.; Yang, Z.; and Nyberg, E.
\newblock 2017.
\newblock Structural embedding of syntactic trees for machine comprehension.

\bibitem[\protect\citeauthoryear{Loper and Bird}{2002}]{Loper2002NLTKTN}
Loper, E., and Bird, S.
\newblock 2002.
\newblock Nltk: The natural language toolkit.
\newblock {\em CoRR} cs.CL/0205028.

\bibitem[\protect\citeauthoryear{Lucas and
  Saccucci}{2000}]{Lucas2000ExponentiallyWM}
Lucas, J.~M., and Saccucci, M.~S.
\newblock 2000.
\newblock Exponentially weighted moving average control schemes: Properties and
  enhancements.

\bibitem[\protect\citeauthoryear{Monsalve, Frank, and
  Vigliocco}{2012}]{Monsalve2012Lexical}
Monsalve, I.~F.; Frank, S.~L.; and Vigliocco, G.
\newblock 2012.
\newblock Lexical surprisal as a general predictor of reading time.
\newblock In {\em Conference of the European Chapter of the Association for
  Computational Linguistics},  398--408.

\bibitem[\protect\citeauthoryear{Nguyen \bgroup et al\mbox.\egroup
  }{2016}]{Nguyen2016MS}
Nguyen, T.; Rosenberg, M.; Song, X.; Gao, J.; Tiwary, S.; Majumder, R.; and
  Deng, L.
\newblock 2016.
\newblock Ms marco: A human generated machine reading comprehension dataset.

\bibitem[\protect\citeauthoryear{Pan \bgroup et al\mbox.\egroup
  }{2017}]{Pan2017MEMEN}
Pan, B.; Li, H.; Zhao, Z.; Cao, B.; Cai, D.; and He, X.
\newblock 2017.
\newblock Memen: Multi-layer embedding with memory networks for machine
  comprehension.

\bibitem[\protect\citeauthoryear{Pennington, Socher, and
  Manning}{2014}]{Pennington2014GloveGV}
Pennington, J.; Socher, R.; and Manning, C.~D.
\newblock 2014.
\newblock Glove: Global vectors for word representation.
\newblock In {\em EMNLP}.

\bibitem[\protect\citeauthoryear{Rajpurkar \bgroup et al\mbox.\egroup
  }{2016}]{rajpurkar2016squad}
Rajpurkar, P.; Zhang, J.; Lopyrev, K.; and Liang, P.
\newblock 2016.
\newblock Squad: 100,000+ questions for machine comprehension of text.
\newblock {\em arXiv preprint arXiv:1606.05250}.

\bibitem[\protect\citeauthoryear{Richardson, Burges, and
  Renshaw}{2013}]{Richardson2013MCTestAC}
Richardson, M.; Burges, C. J.~C.; and Renshaw, E.
\newblock 2013.
\newblock Mctest: A challenge dataset for the open-domain machine comprehension
  of text.
\newblock In {\em EMNLP}.

\bibitem[\protect\citeauthoryear{Seo \bgroup et al\mbox.\egroup
  }{2016}]{seo2016bidirectional}
Seo, M.; Kembhavi, A.; Farhadi, A.; and Hajishirzi, H.
\newblock 2016.
\newblock Bidirectional attention flow for machine comprehension.
\newblock {\em arXiv preprint arXiv:1611.01603}.

\bibitem[\protect\citeauthoryear{Shen \bgroup et al\mbox.\egroup
  }{2017}]{shen2017reasonet}
Shen, Y.; Huang, P.-S.; Gao, J.; and Chen, W.
\newblock 2017.
\newblock Reasonet: Learning to stop reading in machine comprehension.
\newblock In {\em Proceedings of the 23rd ACM SIGKDD International Conference
  on Knowledge Discovery and Data Mining},  1047--1055.
\newblock ACM.

\bibitem[\protect\citeauthoryear{Sordoni \bgroup et al\mbox.\egroup
  }{2016}]{Sordoni2016Iterative}
Sordoni, A.; Bachman, P.; Trischler, A.; and Bengio, Y.
\newblock 2016.
\newblock Iterative alternating neural attention for machine reading.
\newblock {\em arXiv preprint arXiv:1606.02245}.

\bibitem[\protect\citeauthoryear{Srivastava \bgroup et al\mbox.\egroup
  }{2014}]{Srivastava2014Dropout}
Srivastava, N.; Hinton, G.; Krizhevsky, A.; Sutskever, I.; and Salakhutdinov,
  R.
\newblock 2014.
\newblock Dropout: a simple way to prevent neural networks from overfitting.
\newblock {\em Journal of Machine Learning Research} 15(1):1929--1958.

\bibitem[\protect\citeauthoryear{Vinyals, Fortunato, and
  Jaitly}{2015}]{Vinyals2015PointerN}
Vinyals, O.; Fortunato, M.; and Jaitly, N.
\newblock 2015.
\newblock Pointer networks.
\newblock In {\em NIPS}.

\bibitem[\protect\citeauthoryear{Wang and Jiang}{2016}]{Wang2016Machine}
Wang, S., and Jiang, J.
\newblock 2016.
\newblock Machine comprehension using match-lstm and answer pointer.

\bibitem[\protect\citeauthoryear{Wang \bgroup et al\mbox.\egroup
  }{2017}]{Wang2017GatedSN}
Wang, W.; Yang, N.; Wei, F.; Chang, B.; and Zhou, M.
\newblock 2017.
\newblock Gated self-matching networks for reading comprehension and question
  answering.
\newblock In {\em ACL}.

\bibitem[\protect\citeauthoryear{Xiong, Zhong, and
  Socher}{2016}]{xiong2016dynamic}
Xiong, C.; Zhong, V.; and Socher, R.
\newblock 2016.
\newblock Dynamic coattention networks for question answering.
\newblock {\em arXiv preprint arXiv:1611.01604}.

\bibitem[\protect\citeauthoryear{Yang \bgroup et al\mbox.\egroup
  }{2016}]{Yang2016Words}
Yang, Z.; Dhingra, B.; Yuan, Y.; Hu, J.; Cohen, W.~W.; and Salakhutdinov, R.
\newblock 2016.
\newblock Words or characters? fine-grained gating for reading comprehension.

\bibitem[\protect\citeauthoryear{Yu \bgroup et al\mbox.\egroup
  }{2016}]{Yu2016End}
Yu, Y.; Zhang, W.; Hasan, K.; Yu, M.; Xiang, B.; and Zhou, B.
\newblock 2016.
\newblock End-to-end reading comprehension with dynamic answer chunk ranking.

\bibitem[\protect\citeauthoryear{Zeiler}{2012}]{Zeiler2012ADADELTA}
Zeiler, M.~D.
\newblock 2012.
\newblock Adadelta: An adaptive learning rate method.
\newblock {\em Computer Science}.

\end{thebibliography}
\end{document}